\documentclass[sigconf]{acmart}

\AtBeginDocument{%
  }

\setcopyright{acmcopyright}
\copyrightyear{2018}
\acmYear{2018}
\acmDOI{XXXXXXX.XXXXXXX}

\acmConference[Conference acronym 'XX]{Make sure to enter the correct
  conference title from your rights confirmation emai}{June 03--05,
  2018}{Woodstock, NY}
\acmPrice{15.00}
\acmISBN{978-1-4503-XXXX-X/18/06}

\usepackage{microtype}
\usepackage{graphicx}
\usepackage{subfigure}
\usepackage{booktabs} %

\usepackage{amsmath}
\usepackage{enumerate}
\usepackage{enumitem}
\usepackage{hyperref}

\usepackage{algorithm}
\usepackage{algpseudocode}

\usepackage{array}
\usepackage{makecell}

\usepackage{multirow,float}
\usepackage{bm}   
\usepackage{xcolor}
\usepackage{pifont}

\usepackage{tikz,pgfplots}
\usetikzlibrary{pgfplots.groupplots}
\usetikzlibrary{patterns}
\usetikzlibrary{shapes,snakes}
\usetikzlibrary{matrix}
\usetikzlibrary{arrows.meta}
\usepgfplotslibrary{fillbetween}

\usetikzlibrary{external}
\renewcommand{\mathbf}{\bm}

\begin{document}

\title{DistTGL: Distributed Memory-Based Temporal Graph Neural Network Training}

\author{Hongkuan Zhou}
\authornote{The work was performed during an internship at AWS AI Lab.}
\affiliation{%
  \institution{University of Southern California}
  \city{Los Angeles}
  \state{California}
  \country{USA}
}
\email{hongkuaz@usc.edu}

\author{Da Zheng}
\affiliation{%
  \institution{AWS AI}
  \city{Santa Clara}
  \state{California}
  \country{USA}
}
\email{dzzhen@amazon.com}

\author{Xiang Song}
\affiliation{%
  \institution{AWS AI}
  \city{Santa Clara}
  \state{California}
  \country{USA}
}
\email{xiangsx@amazon.com}

\author{George Karypis}
\affiliation{%
  \institution{AWS AI}
  \city{Santa Clara}
  \state{California}
  \country{USA}
}
\email{gkarypis@amazon.com}

\author{Viktor Prasanna}
\affiliation{%
  \institution{University of Southern California}
  \city{Los Angeles}
  \state{California}
  \country{USA}
}
\email{prasanna@usc.edu}

\begin{abstract}
Memory-based Temporal Graph Neural Networks are powerful tools in dynamic graph representation learning and have demonstrated superior performance in many real-world applications.
However, their node memory favors smaller batch sizes to capture more dependencies in graph events and needs to be maintained synchronously across all trainers.
As a result, existing frameworks suffer from accuracy loss when scaling to multiple GPUs. Even worse, the tremendous overhead to synchronize the node memory make it impractical to be deployed to distributed GPU clusters. 
In this work, we propose DistTGL --- an efficient and scalable solution to train memory-based TGNNs on distributed GPU clusters.
DistTGL has three improvements over existing solutions: an enhanced TGNN model, a novel training algorithm, and an optimized system.
In experiments, DistTGL achieves near-linear convergence speedup, outperforming state-of-the-art single-machine method by 14.5\% in accuracy and 10.17$\times$ in training throughput.

\end{abstract}



\maketitle

\section{Introduction}

Recently, along with the success of Graph Neural Networks (GNNs) in static graph representation learning, researchers have designed Temporal Graph Neural Networks (TGNNs)~\cite{tgat,tgn,evolvegcn,dyrep,dysat,jodie} to exploit temporal information in dynamic graphs which is common and important in many real-world applications. For example, in recommender systems, user interests and global trends both change with time. In fraud detectors, the time between two consecutive transactions often marks out suspicious activities. 
In spatial-temporal applications such as traffic and weather prediction, the temporal and spatial information is equally important. 
On various dynamic graphs including social network graphs, traffic graphs, and knowledge graphs, TGNNs have demonstrated superior accuracy on various downstream tasks such as temporal link prediction and dynamic node classification, substantially outperforming static GNNs and other traditional methods~\cite{tgn,tgat}. Depending on whether the timestamps of graph events are discrete or continuous, dynamic graphs can be classified into Discrete Time Dynamic Graphs (DTDGs) and Continuous Time Dynamic Graphs (CTDGs). In this work, we focus on the more general and challenging TGNNs on CTDGs.

\begin{figure}[b]
    \centering
    \input{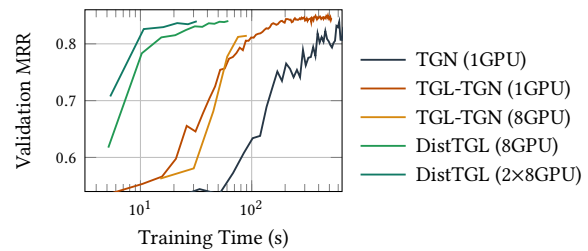}
    \caption{Convergence rate comparison of TGN, TGN implemented using TGL, and DistTGL.}
    \label{fig:teaser}
\end{figure}

On dynamic graphs, the number of related events on each node increases as time evolves. When this number is large, neither temporal attention-based aggregators nor historical neighbor sampling methods allow TGNNs to capture the entire temporal information. To compensate for the lost temporal information, researchers have designed Memory-based Temporal Graph Neural Networks (M-TGNNs)~\cite{jodie,dyrep,tgn,apan} that maintain node-level memory vectors to summarize independent node history. The node memory in M-TGNNs not only allows the aggregator to gather information from fewer historical neighbors but also enlarges the receptive field, as the node memory vectors already contain information multiple hops away. As a result, state-of-the-art M-TGNN TGN~\cite{tgn} only requires a single GNN layer with some recent neighbors as supporting nodes. In the benchmark in TGL~\cite{tgl}, M-TGNNs fill out the top ranks both in accuracy and training time. 

\begin{figure}[t]
    \centering
    \input{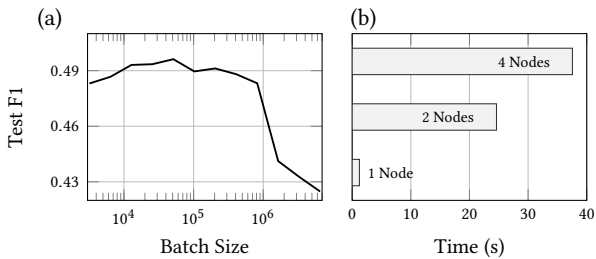}
    \caption{(a) Test accuracy of the GDELT dataset under different batch sizes. (b) Time per epoch spend in reading and writing of the node memory on different numbers of machines. 
    }
    \label{fig:motivation}
\end{figure}

Despite the success of M-TGNNs, it is hard to deploy them to large-scale production applications due to their poor scalability. 
The auxiliary node memory creates temporal dependencies and requires the training mini-batches to be small and scheduled in chronological order. 
Specifically, there are two major challenges to exploiting data parallelism in M-TGNN training.
First, simply increasing the batch size reduces the number of graph events captured in the dynamic node embeddings and leads to information loss (please refer to Section~\ref{sec:batchtraining} for more details).
Figure~\ref{fig:motivation}(a) shows that the accuracy decreases as the batch size increases on the GDELT dataset. On smaller datasets, this decrease in accuracy is usually observed for much smaller batch sizes around $10^3$-$10^4$~\cite{tgn}, where multiple-GPU data parallelism would not provide any throughput improvement over a single GPU. 
Second, all the trainers need to access and maintain a unified version of node memory, making it hard to be deployed to distributed systems.
Unlike static GNN training, these memory operations to the node memory (typically hundreds of megabytes per mini-batch) have strict temporal dependencies.
Due to these excess remote memory operations, distributed systems achieve worse performance than single machines.
Figure~\ref{fig:motivation}(b) shows the case when the node memory is distributed to all machines where each machine owns a unique equally-sized portion. 
Furthermore, the remedy to cross-machine traffics in static GNN training~\cite{distdglv2,zheng2020distdgl,10.1145/3447786.3456233} --- graph partitioning technique METIS~\cite{metis}, is not applicable to dynamic graphs.
As a result, on both small- and large-scale datasets, the training time of the state-of-the-art M-TGNN framework~\cite{tgl} using 8 GPUs on a single node is $10-100\times$ slower than state-of-the-art distributed static GNNs~\cite{distdglv2,bytegnn}, with an unsatisfactory 2-3$\times$ speedup over a single GPU.

In this work, we propose DistTGL --- an efficient and scalable solution to train M-TGNNs on distributed GPU clusters.
DistTGL improves the existing M-TGNN training solutions from three perspectives:
\begin{itemize}
    \item \textbf{Model}: We enhance the node memory in M-TGNNs by adding additional static node memory, which improves both the accuracy and convergence rate.
    \item \textbf{Algorithm}: We design a novel training algorithm to overcome the challenges of accuracy loss and communication overhead in distributed scenarios.
    \item \textbf{System}: We build an optimized system adopting prefetching and pipelining techniques to minimize the mini-batch generation overhead.
\end{itemize}
Compared with existing methods, DistTGL has significant improvement in convergence and training throughput. To the best of our knowledge, DistTGL is the first work that scales M-TGNN training to distributed GPU clusters. DistTGL is publicaly available at Github\footnote{\url{https://github.com/amazon-science/disttgl}}. Our main contributions are
\begin{itemize}
    \item Based on the unique characteristics of M-TGNN training, we propose two novel parallel training strategies --- epoch parallelism and memory parallelism, which allow M-TGNNs to capture the same number of dependent graph events on multiple GPUs as on a single GPU.
    \item We provide heuristic guidelines to determine the optimal training configurations based on the dataset and hardware characteristics.
    \item To overlap mini-batch generation and GPU training, we serialize the memory operations on the node memory and efficiently execute them by an independent daemon process, avoiding complex and expensive synchronizations.
    \item In experiments, DistTGL achieves near-linear speedup when scaling to multiple GPUs in convergence rate, outperforming state-of-the-art single machine method~\cite{tgl} by more than 10$\times$ (see Figure~\ref{fig:teaser}).
\end{itemize}

\section{Background}

Given a dynamic graph, TGNNs aim at embedding the contextual, structural, and temporal information of a given node at a given timestamp into a low-dimensional vector. 
M-TGNNs rely on the node memory and temporal graph attention to generate these vectors. 
We first explain the basic propagation rules in M-TGNNs.
For the rest of this paper, unless stated otherwise, we denote scalar as lower case letter $x$, vector as bold lower case letter $\mathbf x$, and matrix as bold upper case letter $\mathbf X$. We denote row-wise concatenation of vectors (or matrices) using double vertical bar within curly brackets $\{\mathbf x||\mathbf y\}$.

\subsection{Memory-Based Temporal Graph Neural Networks}
M-TGNNs~\cite{jodie,dyrep,tgn,apan} maintain dynamic node-level vectors to track the node history. TGN~\cite{tgn} proposes a general framework for different M-TGNN variants and supports different types of graph events. Here, we introduce TGN on the most common dynamic graphs with graph events of edges appearing. 
For a dynamic graph $\mathcal{G}(\mathcal{V},\mathcal{E})$, its graph events could be represented by a time-ordered series $\{(u, v, \mathbf{e}_{uv}, t)\}$ where each quadruple represents an edge with edge feature $\mathbf{e}_{uv}$ occurring between node $u$ and node $v$ at time $t$.
For each node $v\in\mathcal{V}$, we maintain a node memory vector $\mathbf{s}_v$, which is initialized to be a zero vector. When an edge $e_{uv}$ connecting node $u$ and node $v$ appears at timestamp $t$, two mails are generated at node $u$ and node $v$
\begin{align}
    \mathbf m_u&=\left\{\mathbf s_u||\mathbf s_v||\Phi(t-t^-_u)||\mathbf e_{uv}\right\}\\
    \mathbf m_v&=\left\{\mathbf s_v||\mathbf s_u||\Phi(t-t^-_v)||\mathbf e_{uv}\right\},
\end{align}
where $\Phi(\cdot)$ is the time encoding~\cite{tgat}, $t^-_u$ is the timestamp when $\mathbf s_u$ is last updated, and $\mathbf e_{uv}$ is the edge feature. Then, we use an update function $\textup{UPDT}$ to update the node memory of node $u$ and node $v$,
\begin{align}
    \mathbf s_u=\textup{UPDT}\left(\mathbf s_u,\mathbf m_u\right)\qquad \mathbf s_v=\textup{UPDT}\left(\mathbf s_v,\mathbf m_v\right)\label{eq:memupdt}
\end{align}
The update function can be implemented using any sequence model. In TGN-attn~\cite{tgn}, $\textup{UPDT}(\cdot)$ is implemented as GRU cells.
Since the $\textup{UPDT}$ function is only called when a related graph event occurs, the lengths of the hidden status of different nodes in the graph are different. 
In backward propagation, the learnable parameters $\bm W$ and $\bm b$ are trained within each GRU cell (the gradients do not flow back to previous GRU cells, like in the Back-Propagation-Through-Time algorithm).

After updating the node memory, a one-layer temporal attention layer~\cite{tgat} gathers and aggregates information from the node memory of the most recent neighbors $\mathbf S_w,w\in\mathcal N_v$ to compute the dynamic node embedding $\mathbf h_v$ for node $v$. If dynamic or static node features are available, they can be combined with the node memory.
\begin{align}
    \mathbf q&=\mathbf W_q\{\mathbf s_v||\mathbf\Phi(0)\}+\mathbf b_s \label{eq:attq}\\
    \mathbf K&=\mathbf W_k\{\mathbf S_w||\mathbf E_{vw}||\mathbf\Phi(\Delta \mathbf t)\}+\mathbf b_k \label{eq:attk}\\
    \mathbf V&=\mathbf W_v\{\mathbf S_w||\mathbf E_{vw}||\mathbf\Phi(\Delta \mathbf t)\}+\mathbf b_v \label{eq:attv}\\
    \mathbf h_v&=\textup{Softmax}\left(\frac{\mathbf q\mathbf K^\textup{T}}{\sqrt{|\mathcal N_v|}}\right)\mathbf V \label{eq:att},
\end{align}
where $\Delta \mathbf t$ is the time differences of the current timestamp with the last updated time of the node memory of $w\in\mathcal N_v$, and $\mathbf E_{vw}$ is the matrix of edge features connecting nodes $v$ and $w\in\mathcal N_v$.

Most TGNNs are self-supervised using the temporal edges as ground truth information, where the updates to node memory are delayed by one iteration due to the information leak problem~\cite{tgn}. Specifically, the mails are cached for the supporting nodes, and the output embeddings are computed using Equation \ref{eq:attq}-\ref{eq:att} before their node memory is updated using Equation \ref{eq:memupdt}. This reversed computation order needs to be implemented both in training and at inference to avoid the information leak problem.

\subsubsection{Batched M-TGNN Training}
\label{sec:batchtraining}

\begin{figure}[t]
  \centering
  \input{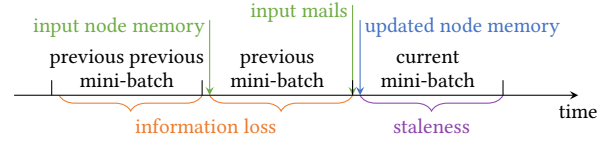}
  \caption{Overview of the inaccuracy in node memory caused by batched training.}
  \label{fig:nodememory}
\end{figure}

Since the training of M-TGNNs needs to be synchronized with the node memory, the training samples need to be scheduled chronologically.
Theoretically, the node memory of a node needs to be immediately updated after a relevant graph event occurs on that node so that later dependent nodes can use this up-to-date node memory in the message passing process.
Without changing the algorithm, we can process consecutive graph events that do not have overlapping nodes in batches by updating their node memory in parallel. 
However, this limits the batch size to no more than a few graph events on most dynamic graphs.
In practice, the tiny batch size is computationally infeasible on modern hardware intended for highly paralleled programs. 
To solve this problem, M-TGNNs process the incoming graph events in larger fixed-size batches and update the node memory for the nodes that have new mails once per batch to reduce the computation time.
Let $\{\mathbf m_u\}$ be the set of mails generated at node $u$ in a batch of graph events, $\mathbf s_u$ is then updated using a $\textup{COMB}(\cdot)$ function
\begin{equation}
    \mathbf s_u= \textup{UPDT}(\mathbf s_u,\textup{COMB}(\{\mathbf m_u\})).
\end{equation}
Note that the mails $\{\mathbf m_u\}$ is not using the up-to-date node memory (since it is not computed yet) but using the outdated node memory at the last batch of graph events.
In TGN-attn, the $\textup{COMB}(\cdot)$ function simply outputs the most recent mail.
This batching approach both updates the node memory in batch and computes the attention-based message passing in batch.
The batched update to node memory causes two types of inaccuracy in the node memory --- staleness and information loss (Figure~\ref{fig:nodememory}).
The staleness in the node memory refers to the problem where the node memory is not up-to-date due to the reversed computation order to avoid the information leak problem.
The information loss in the node memory refers to the node memory not being updated by the mails that are filtered out by the $\textup{COMB}(\cdot)$ function as well as the inaccuracy of the mails due to using outdated node memory.
When the batch size is increased, both the staleness and information loss in the node memory increase, resulting in lower accuracy~\cite{tgn}. Besides these two types of inaccuracy, another common inaccuracy in sequence models is the inaccuracy due to not re-computing the hidden embeddings when the weights are updated, which generally does not affect the performance.

\begin{figure*}[t]
  \centering
  \input{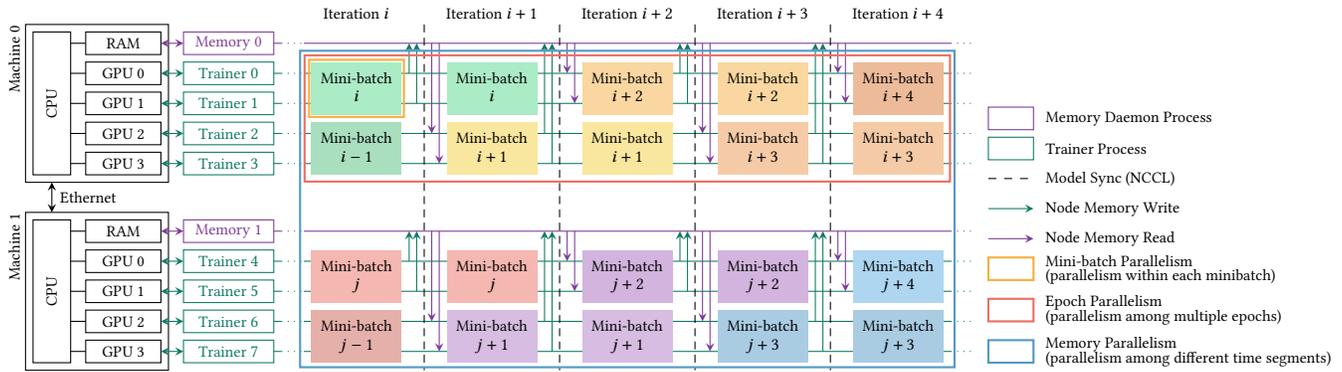}
  \caption{Overview of DistTGL training with $2\times2\times2$ (mini-batch$\times$epoch$\times$memory) parallelism on two four-GPU machines. For simplicity and easier understanding, we draw the reads and writes to the node memory at the beginning and end of each training iteration. In our optimized system, they have performed asynchronously with the training iterations and are fully overlapped with the GPU computation. Please refer to Figure~\ref{fig:3paral} for more details on the three parallel training strategies.}
  \label{fig:main}
\end{figure*}

\subsection{Related Works}

Dynamic graph representation learning plays an important role in many real-world problems. Many discrete TGNNs~\cite{evolvegcn,dysat,goyal2018dynamicgem,hajiramezanali2019variational}, continuous TGNNs~\cite{dyrep,tgat,tgn,apan}, and non-GNN methods~\cite{knowe,cawn} are proposed to learn node embeddings on dynamic graphs. 
There are many existing works that accelerate the message passing scheme in GNNs on a single node~\cite{dgl,Fey/Lenssen/2019} and on distributed GPU clusters~\cite{zheng2020distdgl,distdglv2,10.1145/3447786.3456233,bytegnn,euler}.
In discrete TGNNs, the propagation within a graph snapshot is the same as static GNNs where these existing methods can be directly applied to.
There are also some existing works that specialize in discrete TGNNs on a single GPU~\cite{you2022roland,10.1145/3534678.3539352} and distributed systems~\cite{chakaravarthy2021efficient}. 
However, these methods do not apply to continuous M-TGNNs due to the unique propagation rule of M-TGNNs.
Accelerating continuous M-TGNNs is challenging due to the aforementioned antithesis between training speed and accuracy. 
Distributed M-TGNN training is even more challenging due to the high volume of data synchronization.
There are a few works that accelerate M-TGNNs training.
TGL~\cite{tgl} proposes a general framework for single-node multiple-GPU continuous TGNNs. However, TGL does not support distributed GPU clusters. The speedup of TGL on multiple GPUs in a single machine is also unsatisfactory, only achieving $2-3\times$ speedup on 8 GPUs. 
EDGE~\cite{chen2021efficient} proposes to speedup the training by replacing the dynamic node memory of active nodes with static learnable node memory, gambling on the chance that active nodes have stable embeddings. 
To the best of our knowledge, there is no existing work for M-TGNN training that achieves near-linear scalability on single-node multiple-GPU, or operates on distributed GPU clusters.
For the inference task, TGOpt~\cite{Wang2023TGOptRO} proposes to accelerate TGNN inference by de-duplication, memorization, and pre-computation. Another work~\cite{9820671} proposes a system-architecture co-design that accelerates M-TGNN inference on FPGAs. Unfortunately, these techniques do not apply to M-TGNN training.
\section{DistTGL}

\begin{figure}[b]
  \centering
  \input{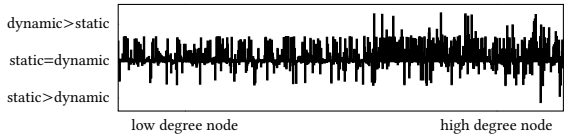}
  \caption{Accuracy differences of each node with static and dynamic node memory on the Wikipedia dataset, sorted by node degrees. Positive bars (in dynamic$>$static region) indicate that dynamic node memory has better accuracy than static node memory for those nodes, and vice versa.}
  \label{fig:memdiff} 
\end{figure}

We propose DistTGL --- an efficient and scalable solution to train M-TGNNs on distributed GPU clusters. DistTGL achieves scalability through improvements from three perspectives: model, algorithm, and system. From the model perspective, we introduce the static node memory that explicitly separates the time irrelevant node information. From the algorithm perspective, we propose two novel parallel training strategies and a method to determine the best combination of these strategies on any given dataset and hardware configuration. From the system perspective, we design an efficient system to reduce and overlap mini-batch generation overhead with GPU training. We introduce these improvements in the three following subsections.

\subsection{M-TGNN Model with Static Node Memory}
\label{sec:disttglmodel}

M-TGNNs rely on node memory to summarize the node history. Previous work~\cite{chen2021efficient} argues that the node memory of nodes with active interactions is static. While this may be true on some evolving graphs like citation graphs, it fails on the dynamic graphs where high-frequency information is important, such as in fraud detection~\cite{siering2017taxonomy}. Figure~\ref{fig:memdiff} shows the comparison of the accuracy in the temporal link prediction task that predicts destination nodes from source nodes using static and dynamic node memory. We do not observe any noticeable inclination on higher degree nodes favors static node memory or vice versa. We also observe similar results on the other datasets used in this work.

\begin{figure}[b]
  \centering
  \input{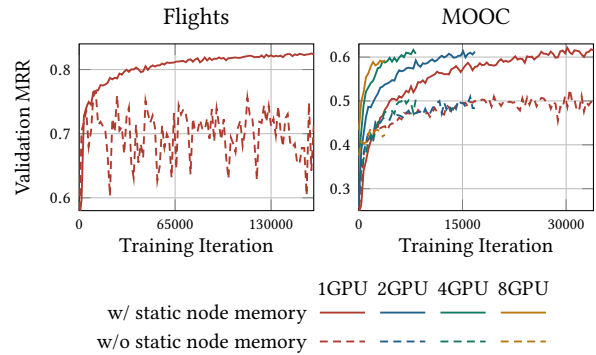}
  \caption{Validation accuracy on the Flights and MOOC datasets with and without pre-trained static node memory.}
  \label{fig:modelimprov.tex}
\end{figure}

\begin{figure*}[t]
  \centering
  \input{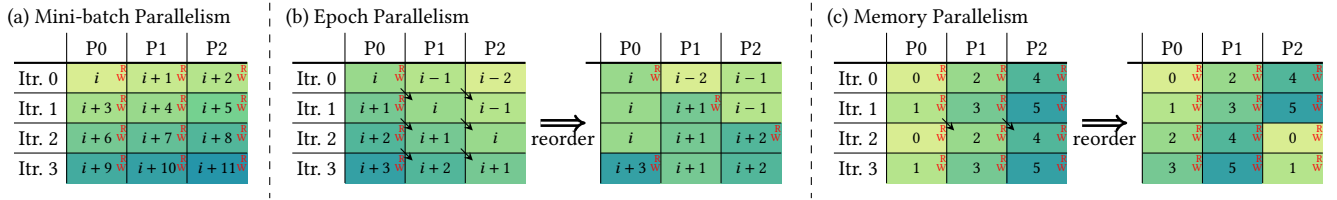}
  \caption{Overview of mini-batch parallelism, epoch parallelism, and memory parallelism on three trainer processes. The ``R'' and ``W'' denote read and write operations to the shared node memory. In epoch parallelism, the arrows denote cross-process communication to send mini-batch data. In memory parallelism, the arrows denote cross-process communication to send the updated node memory. }
  \label{fig:3paral}
\end{figure*}

We believe that a general TGNN model should be able to capture both the dynamic and static node information of all nodes. In DistTGL, we separate the static and dynamic node memory and capture them explicitly. DistTGL keeps the original GRU node memory on all nodes to capture the dynamic node information and implements an additional mechanism to capture the static node information. There are two major benefits brought by this additional static node history. First, it enhances the capability of M-TGNNs to capture node history with burst interactions.
Due to the batching of updating the node memory, if a node interacts with others many times in a short time period, it is inevitable that the $\textup{COMB}(\cdot)$ function used in the dynamic node memory would filter out most of these interactions, resulting in a loss of high-frequency information.
The static node memory, combined with the time encoding~\cite{tgat} in the temporal attention aggregator, could boost the performance in such cases.
Second, the static node memory explicitly separates the information irrelevant to batch sizes, which improves the performance of data parallelized training.
Since the static node memory is irrelevant with time, all graph events can be used to supervise the training process, allowing it to capture all static information regardless of batching.
In this work, since most dynamic graphs do not have node features, we use learnable node embeddings pre-trained with the same task as the static node memory due to its simplicity. 
The pre-training of these embeddings can be easily done in any well-optimized distributed static GNN frameworks~\cite{zheng2020distdgl,distdglv2,10.1145/3447786.3456233,bytegnn,euler}.
Note that the static node memory is similar to learnable weights in the M-TGNN models and does not include any information in the test set. On the other hand, the dynamic node memory contains information in the test set and would cause information leaks if not handled properly.
DistTGL also supports other kinds of learnable or non-learnable static node memory, such as co-trained embedding tables or even node embeddings generated by static GNNs.

Figure \ref{fig:modelimprov.tex} shows the two datasets which have the most significant improvement with pre-trained static node memory. On a single GPU, our improved model achieves remarkably better accuracy on both datasets and a smoother convergence curve on the Flights dataset (we do not show the curves for multi-GPU for a clearer visualization). On the MOOC dataset, our model with static node memory also improves the scalability in convergence on multiple-GPU using epoch parallelism (which will be introduced later in Section \ref{sec:disttglalgo}).

\subsection{Parallel Training Algorithm}
\label{sec:disttglalgo}

A straightforward approach to train M-TGNNs in parallel is to process the graph events in large global batches and distribute them to multiple trainers, which is used by TGL~\cite{tgl} in the setting of multiple GPUs on a single node. We refer to this approach as the mini-batch parallelism, which relaxes the inter-batch dependencies in node memory.
However, the key to achieving good accuracy in multi-GPU M-TGNN training is to maintain the temporal dependency when the graph events are processed in large batches. 
To solve this problem, we propose two novel parallel training strategies --- epoch parallelism and memory parallelism.
Epoch parallelism relaxes the dependencies in the node memory due to weight updates and trains different epochs simultaneously on different trainers.
Memory parallelism trades space for accuracy by maintaining multiple copies of the node memory at different timestamps.
In the rest of this section, we first introduce the three types of parallelism and their advantages and disadvantages. 
Then, we discuss how to design an optimal training algorithm given any task specifications and hardware configurations.

\subsubsection{Mini-Batch Parallelism} 
Mini-batch parallelism simply trains a large global batch on multiple trainers in parallel. On $n$ GPUs, a global batch of graph events is evenly divided into $n$ local batches where each GPU is responsible for computing the output embeddings of one local batch. 
Figure~\ref{fig:3paral}(a) shows the case when a global batch is divided into three local batches on three trainers.
Since the global mini-batches are generated in chronological order, we also split them into local mini-batches chronologically and ignore the intra-dependency within each global mini-batch. Specifically, these $n$ trainers first fetch the node memory and cached mails of the assigned root nodes and their supporting nodes. Then, they compute the forward and backward propagation and update the model weights. Before they use the computed node memory to update the node memory and cached mails, they need to make sure all trainers have finished the fetch operations to avoid Write-After-Read (WAR) hazard. Note that ideally, the node memory and cached mails should be updated for both the root and supporting nodes so that we do not need to re-compute Equation~\ref{eq:memupdt} when these supporting nodes are referenced again in later batches. However, to ensure the model weights can receive enough feedback in the backward propagation, we do not update the node memory and cached mails of the supporting nodes and re-compute them when they are referenced later. Because the fetch and update of the node memory are done simultaneously in all trainers, the node embeddings generated for later graph events in the global batch cannot perceive the earlier graph events, incurring both staleness and information loss in the node memory. In addition, mini-batch parallelism requires all trainers to maintain the same copy of node memory, which leads to enormous communication overhead on distributed systems.

\subsubsection{Epoch Parallelism} Epoch parallelism leverages data parallelism by training different epochs simultaneously using only one copy of the node memory. In the vanilla M-TGNN training, self-supervised by temporal edges on a single GPU, we first sample some negative destination nodes for the root nodes in mini-batch $i$. We then collect the supporting nodes for all positive and negative root nodes and fetch their node memory and cached mails. In the later epochs, for the same root nodes in mini-batch $i$, we sample different sets of negative destination nodes and follow the same procedure to get their node memory and cached mails. To train on the same mini-batches in different epochs in parallel on $n$ trainers, we ignore the difference in node memory due to weight updates in the last $n-1$ epochs. Thus, we can prepare one set of inputs of the positive nodes and $n$ sets of inputs of the negative nodes and train them in parallel. Note that these mini-batches need to be scheduled in different iterations so that the gradients of positive nodes are not simply multiplied by $n$. This scheduling increases the variance of the gradients of the sampled mini-batches, as the same set of positive nodes is learned for $n$ consecutive iterations. The left part of Figure~\ref{fig:3paral}(b) shows the case when applying epoch parallelism to three trainers. In each iteration, trainer P0 fetches the node memory and cached mails for one positive mini-batch and three negative mini-batches. After P0 finishes one iteration, it writes to the node memory and sends the prepared mini-batches (one positive mini-batch and the two unused negative mini-batches) to P1. P1 receives the mini-batches from P0 and sends them (one positive mini-batch and the unused one negative mini-batch) to P2 after the computation. Note that only P0 needs to write back the updated node memory to the global copy of node memory in the main memory. Although the node memory of this mini-batch in P1 and P2 is updated using a more recent version of the weights, writing them to the global copy would lead to Read-After-Write (RAW) hazards with later training iterations. We also tried a finer-grained updating policy which updates nodes that do not have this RAW hazard in P1 and P2. However, it does not outperform the original policy. To reduce the cross-trainer communication, we further optimize the algorithm by reordering the mini-bathes so that each trainer works on the same positive samples (with different negative samples) for $n$ consecutive iterations (see the right part in Figure~\ref{fig:3paral}(b)). However, epoch parallelism still requires all trainers to access the same node memory, which is impractical on distributed systems.

\subsubsection{Memory Parallelism} Memory parallelism trades space for time by training different time segments of the dynamic graph simultaneously using separate copies of node memory. The left part in Figure~\ref{fig:3paral}(c) shows the case when applying memory parallelism on a dynamic graph with 6 mini-batches with three trainers and three copies of node memory. Each trainer is only responsible for one-third of the whole dynamic graph, i.e., a time segment of two consecutive mini-batches. In every iteration, each trainer needs to fetch its own node memory and cached mails. The design on the left requires the intermediate node memory to be transferred across the processes after the trainers finish their time segments. For example, P0 needs to send the node memory of all the nodes in the graph to P1 after iteration 1, which is expensive in distributed systems. To solve this problem, we reorder the mini-batches across the trainer (see the right part in Figure~\ref{fig:3paral}(c)) so that each trainer trains sequentially on all the segments using its own node memory. Since each trainer owns its individual node memory, there is no synchronization of the node memory across the trainers, making it the only suitable strategy for distributed systems.

\begin{table}[t]
    \setlength{\tabcolsep}{0.9mm}
    \centering
    \caption{Summary of the three parallel training strategies on $n$ trainers. The comparison with single-GPU training is made based on the same local batch size. The ``Training overhead'' row refers to the overheads in mini-batch generation at the beginning of each training iteration. The advantages are marked in \textbf{bold}.}
    \begin{tabular}{r||>{\centering\arraybackslash}p{1.8cm}|>{\centering\arraybackslash}p{1.8cm}|>{\centering\arraybackslash}p{1.8cm}}
         & Mini-batch Parallelism & Epoch Parallelism & Memory Parallelism \\
        \toprule
        \makecell{Captured\\dependency} & \makecell{less than\\single-GPU} & \makecell{\textbf{same as}\\\textbf{single-GPU}} & \makecell{\textbf{same as}\\\textbf{single-GPU}} \\
        \midrule
        \makecell{Training\\overhead} & \makecell{\textbf{same as}\\\textbf{single-GPU}} & \makecell{$n$ times\\single-GPU} & \makecell{\textbf{same as}\\\textbf{single-GPU}}\\
        \midrule
        \makecell{Main memory\\requirement} & \makecell{\textbf{same as}\\\textbf{single-GPU}} & \makecell{\textbf{same as}\\\textbf{single-GPU}} & \makecell{$n$ times\\single-GPU}\\
        \midrule
        \makecell{Synchronization\\across trainers} & \makecell{weights and\\node memory} & \makecell{weights and\\node memory} & \makecell{\textbf{weights}\\\textbf{only}}\\
        \midrule
        \makecell{Gradient descent\\variance} & \makecell{\textbf{same as}\\\textbf{single-GPU}} & \makecell{more than\\single-GPU} & \makecell{\textbf{same as}\\\textbf{single-GPU}}\\
    \end{tabular}
    \label{tab:3para}
\end{table}

\subsubsection{Optimal Training Algorithm} 
The aforementioned three parallelization strategies all have their own unique characteristics. We summarize their advantages and disadvantages in Table~\ref{tab:3para}. 
To achieve optimal training performance, we provide heuristic guidelines for DistTGL users to combine these strategies to pick their advantages and offset their disadvantages.
Consider a distributed system with $p$ machines and $q$ GPUs per machine. Let $i\times j\times k=p\times q$ be a training configuration where $i$ represents how many GPUs to compute each mini-batch, $k$ represents how many copies of node memory to maintain, and $j$ represents how many epochs to train in parallel for each copy of node memory. 
We determine the optimal choice of $(i,j,k)$ from task requirements and hardware configurations. There are two constraints from the hardware side. First, we need to have $k\geq p$ as memory parallelism is the only strategy that does not synchronize node memory across the trainers. Then, the main memory of each machine should be able to hold $k/p$ copies of node memory and cached mails, or at least hold sufficient cache if using the disk-based memory caching storage option. 

\begin{figure}[b]
  \centering
  \input{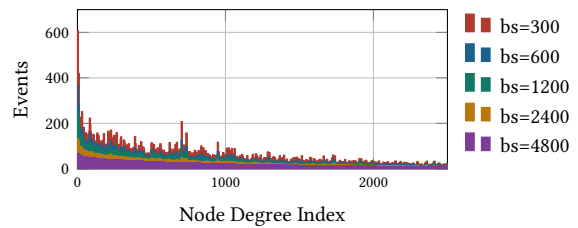}
  \caption{Number of captured events in the node memory with different batch sizes, sorted by node degree from high to low on the Wikipedia dataset.}
  \label{fig:memfreq}
\end{figure}

\begin{figure*}[t]
    \centering
    \input{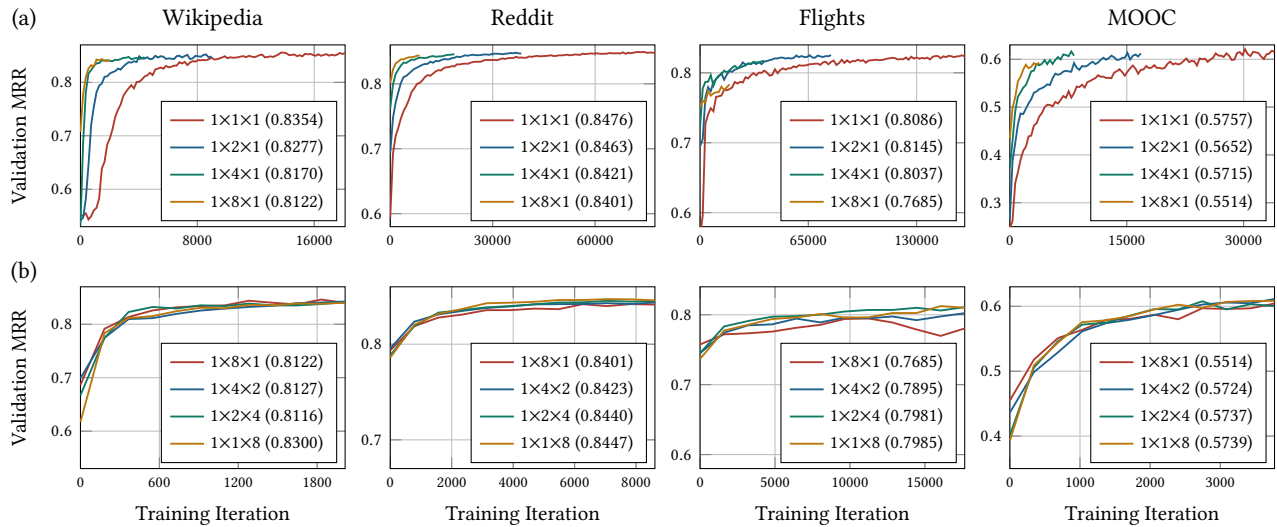}
    \caption{(a) Convergence curve of DistTGL with different epoch parallelism $j$ using 1-8 GPUs on one node. (b) Convergence curve of DistTGL with combination of different epoch and memory parallelism $j\times k$ using 8 GPUs on one node. The test MRR is shown between parentheses in the legend. Compared with the single-GPU baseline, DistTGL with memory parallelism $1\times1\times8$ achieves near-linear converge speedup with negligible accuracy loss on 8 GPUs.}
    \label{fig:epoch+memory}
\end{figure*}

Under these constraints, we first determine $i$ according to the largest batch size. 
Figure~\ref{fig:memfreq} shows that when the batch size increases, fewer graph events would be captured in the node memory, especially for high-degree nodes. DistTGL users can set a threshold for the amount of missing information so that DistTGL would reversely find out the largest batch size. For applications where high-frequency information is crucial, we can set a stricter threshold for high-degree nodes. Based on this batch size, $i$ can be determined according to the GPU specifications. For $j$ and $k$, we always prefer to apply memory parallelism since it leads to better convergence, which we have also verified from experiments (see Figure~\ref{fig:epoch+memory}.(b)). In summary, we first determine $i$ based on task requirements, then $k$ based on hardware specification, and lastly $j$ is fixed by $p\times q/i\times k$.

For example, on a distributed system with 4 machines and 8 GPUs each machine, we determine the largest batch size is 3200 edges. The GPU saturates when batch size is larger than 1600 edges. So we first set local batch size to be 1600 edges and $i=2$. The main memory of each machine can hold two copies of the node memory. Then we set $k=32/2/2=8$. Finally, $j$ is fixed to be $32/2/8=2$.

\subsection{Distributed Training System}
\label{sec:disttglsys}

Designing a scalable distributed training system for M-TGNNs is not trivial. Even for the most straightforward mini-batch parallelism, previous work~\cite{tgl} only achieves 2-3$\times$ speedup using 8 GPUs on a single node due to excessive overheads in mini-batch generation. 
We solve this issue by prefetching the mini-batches in a separate process and pipelining the sub-tasks  (loading from disk, slicing features, slicing node memory, writing back to node memory) within one mini-batch generation.
Figure~\ref{fig:main} shows an overview of DistTGL serializing the memory operations and executing them asynchronously on separate processes. Here we focus on describing the most important design that handles the reads and writes to the node memory. As memory parallelism works on separate copies of node memory which has no dependency and can be easily parallelized, we consider the case for each $i\times j$ trainer group that shares the same copy of the node memory. Since $k\geq p$, each trainer group must have all the processes on the same physical machine. Within each $i\times j$ group, the memory operations can be serialized as a spin lock acting on each $i$ sub-group. For example, for $i\times j=2\times 2$, we have the memory access sequence
\begin{align*}
  (\textup R_0\textup R_1)(\textup W_0 \textup W_1)(\textup R_2\textup R_3)(\textup W_2\textup W_3)(\textup R_0\textup R_1)(\textup W_0 \textup W_1)\cdots,
\end{align*}
where $\textup R_i$ and $\textup W_i$ denote read and write requests from trainer $i$, and there is no ordering for the requests within each bracket. 

\begin{algorithm}[t]
  \caption{Memory Daemon Process}
  \label{alg:mem}
\begin{algorithmic}
  \State {\bfseries Input:} \texttt{read1\_idx\_buf}, \texttt{mem\_write\_buf}, \texttt{mail\_write\_buf}, \texttt{write\_1idx\_buf}
  \State {\bfseries Modify:} \texttt{read\_status}, \texttt{write\_stats}
  \State {\bfseries Output:} \texttt{mem\_read\_buf}, \texttt{mail\_read\_buf}
  \Repeat
    \State reset \texttt{memory} and \texttt{mail}
    \State \texttt{rank} = 0
    \Repeat
      \renewcommand{\algorithmicdo}{\textbf{do in parallel}}
      \For {$r$ in [\texttt{rank},\texttt{rank} + $j$)}
        \State wait until \texttt{write\_status}[$r$] == 1
        \State write to \texttt{memory} from \texttt{mem\_write\_buf}[$r$]
        \State write to \texttt{mail} from \texttt{mail\_write\_buf}[$r$]
        \State \texttt{write\_status}[$r$] = 0
      \EndFor
      \State \texttt{rank} += $i$
      \State \texttt{rank} = 0 if \texttt{rank} == $i\times j$
      \For {$r$ in [\texttt{rank},\texttt{rank} + $j$)}
        \State wait until \texttt{read\_status}[$r$] == 1
        \For {$jj$ in [0, $j$)}
          \State slice \texttt{memory} to \texttt{mem\_read\_buf}[$r$][$jj$]
          \State slice \texttt{mail} to \texttt{mail\_read\_buf}[$r$][$jj$]
        \EndFor
        \State \texttt{read\_status}[$r$] = 0
      \EndFor
      \renewcommand{\algorithmicdo}{\textbf{do}}
    \Until {epoch end}
  \Until {training end}
\end{algorithmic}
\end{algorithm}

In DistTGL, instead of implementing an expensive cross-process lock mechanism, we launch an additional memory daemon process for each group of $i\times j$ trainer processes to handle the read and write requests for all the trainers in that group. Let $bs$ be the local batch size, $d$ be the number of sampled supporting nodes for each root node, and $d_{\textup{mem}}$ be the dimension of the node memory. The memory process allocates the following buffers, which are shared with the trainers:
\begin{itemize}
  \item \texttt{mem\_read\_buf} of size $\left[i\times j,j,bs\times d,d_{\textup{mem}}\right]$ that holds the results of the memory read requests.
  \item \texttt{mail\_read\_buf} of size $\left[i\times j,j,bs\times d,2d_{\textup{mem}}\right]$ that holds the results of the mail read requests.
  \item \texttt{read\_1idx\_buf} of size $\left[i\times j,j,bs\times d+1\right]$ that holds the indexes of the read requests and its length.
  \item \texttt{mem\_write\_buf} of size $\left[i\times j,bs,d_{\textup{mem}}\right]$ that holds the input of the memory write request.
  \item \texttt{mail\_write\_buf} of size $\left[i\times j,bs,2d_{\textup{mem}}\right]$ that holds the input of the mail write request.
  \item \texttt{write\_1idx\_buf} of size $\left[i\times j,bs+1\right]$ that holds the indexes of the read requests and its length.
  \item \texttt{read\_status} of size $\left[i\times j\right]$ that indicates the status of the read request.
  \item \texttt{write\_status} of size $\left[i\times j \right]$ that indicates the status of the write request. 
\end{itemize}
Algorithm~\ref{alg:mem} shows the pseudo-code of the memory daemon process. Each trainer process issues the read and write requests by copying the inputs to the shared buffers and setting the elements of its rank in \texttt{read\_status} and \texttt{write\_status} to be 1. The memory daemon process executes these requests in serialized order, puts the read results to the buffers, and resets the status. Note that the first read request of each epoch is not issued, as the results are always all zero matrices right after the initialization.

\section{Experiments}
\label{sec:exp}

We perform detailed experiments to evaluate the performance of DistTGL. We implement DistTGL using PyTorch~\cite{pytorch} 1.11.0 and DGL~\cite{dgl} 0.8.2. The code and datasets will be open-sourced upon acceptance of this work.

\begin{table}[b]
    \setlength{\tabcolsep}{0.9mm}
    \centering
    \caption{Dataset Statistic. The max$(t)$ column shows the maximum edge timestamp (minimum edge timestamp is 0 in all datasets). $|d_v|$ and $|d_e|$ show the dimensions of node features and edge features, respectively. The * mark denotes pre-trained features.}
    \begin{tabular}{r|ccccccc}
        & $|V|$ & $|E|$ & max$(t)$ & $|d_v|$ & $|d_e|$\\
        \toprule
        Wikipedia & 9,227 & 157,474 & 2.7e6 & 100* & 172\\
        Reddit & 10,984 & 672,447 & 2.7e6 & 100* & 172\\
        MOOC & 7,144 & 411,749 & 2.6e7 & 100* & -\\
        Flights & 13,169 & 1,927,145 & 1.0e7 & 100* & -\\
        GDELT & 16,682 & 191,290,882 & 1.6e8 & 413 & 130\\
    \end{tabular}
    \label{tab:ds}
\end{table}

\noindent \textbf{Datasets.} Table~\ref{tab:ds} shows the statistics of the five datasets for the evaluation. The task on each dataset is
\begin{itemize}
    \item \textbf{Wikipedia}~\cite{10.1145/3292500.3330895} is a bipartite user-internet page graph where one graph event represents one user modifies the one Wikipedia page. The edge features are extracted from the text that the users update the pages with. The task on this dataset is temporal link prediction.
    \item \textbf{Reddit}~\cite{10.1145/3292500.3330895} is a bipartite user-reddit graph where one graph event represents one user posts to one sub-reddit. The edge features are extracted from the text of the post. The task on this dataset is temporal link prediction.
    \item \textbf{MOOC}~\cite{10.1145/3292500.3330895} is a bipartite user-course action graph where one graph event represents one user interacting with one class item (i.e., watching a video, answering a question). The task on this dataset is temporal link prediction.
    \item \textbf{Flights}~\cite{poursafaei2022towards} is a traffic graph where each node represents one airport, and each edge represents one flight between the two airports. The task on this dataset is temporal link prediction.
    \item \textbf{GDELT}~\cite{tgl} is a knowledge graph tracking events happening all over the world where each node represents one actor, and each edge represents one event. Since the temporal link prediction task used in TGL~\cite{tgl} is too simple, we use the 130-dimensional CAMEO code as edge features and set the task to be a 56-class 6-label dynamic edge classification problem that predicts the rest of the 56-dimensional edge features. 
\end{itemize} 
For the temporal link prediction task, to reduce the variance in the validation and test accuracy, we randomly sample 49 negative destination nodes (for bipartite graphs, we only sample from the other graph partition) and report the Mean Reciprocal Rank (MRR) of the true destination nodes. 
For the dynamic edge classification task, we report the F1-Micro score.

\subsubsection{Model} We use the most efficient one-layer TGN-attn~\cite{tgn} model enhanced with the static node memory introduced in Section~\ref{sec:disttglmodel}. We follow the original work to set the dimension of node memory to 100 and the number of most recent neighbors to 10 for each node. We pre-train the static node history with the same GNN architecture but only with static information using DGL~\cite{dgl}. On the Wikipedia, Reddit, MOOC, and Flights datasets, we pre-train 10 epochs with stochastically selected mini-batches. On the GDELT dataset, we only pre-train 1 epoch. The pre-training of all datasets takes less than 30 seconds on a single machine. For the Wikipedia, Reddit, MOOC, and Flights datasets, we set the local batch size to be the largest available batch size 600~\cite{tgl}. For the GDELT dataset, the local batch size is set to 3200, limited by the GPU capacity. We set the learning rate to be linear with the global batch size. To ensure fairness, we keep the total number of traversed edges to be the same in multi-GPU training. The number of training iterations for $x$ GPUs will be $1/x$ compared to a single GPU. On the Wikipedia, Reddit, MOOC, and Flights datasets, we traverse the training events 100 times (100 epochs on a single GPU). On the larger GDELT dataset, we traverse the training events 10 times (10 epochs on a single GPU). On the Wikipedia, Reddit, MOOC, and Flights datasets, we perform evaluation after every training epoch using the node memory in the first memory process. On the GDELT dataset, due to the slow evaluation process (as DistTGL only accelerates training), we perform validation and testing every 2000 training iterations on a randomly selected chunk of 1000 consecutive mini-batches in the validation and the test set, starting with all-zero node memory and mails. 

\begin{figure}[t]
    \centering
    \input{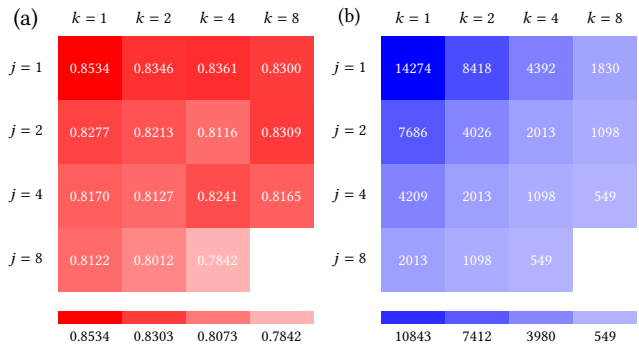}
    \caption{(a) Test MRR and (b) number of iterations before convergence with different epoch parallelism $j$ and memory parallelism $k$ on the Wikipedia dataset.}
    \label{fig:heat}
\end{figure}

\subsubsection{Hardware} All experiments are performed on AWS EC2 cloud using
g4dn.metal instances with dual Intel Platinum 8259CL CPU paired with 384GB EC-DDR4 memory, 8 Nvidia T4 GPUs with 16GB GDDR6 memory for each GPU, two 900GB NVMe SSDs, and 100Gbps Ethernet connection. We create the instances in the same group of rack to make sure the cross-machine latency is minimized. Due to the lack of CPU cores, we sample the mini-batch in advance and store them on the two NVMe SSDs in RAID0 mode to maximize the throughput, which could also be generated on the fly during training if the CPU power is enough. Note that the positive edges in the mini-batches are reused in every epoch. For the negative edges, we observe that in the temporal link prediction task, a small number of groups of negative edges are enough. So we prepare 10 groups of negative edges and randomly use them in the total 100 epochs. We assign 6 CPU threads for each trainer and memory process so that the total 96 physical threads can serve the needs for maximum memory parallelism of $k=8$ on a single machine. To further overlap the mini-batch generation with the GPU computation, we pre-fetch the pre-sampled static information from disks $j$ iterations in advance. We directly pre-fetch the static information to GPU using a separate CUDA stream than the training CUDA stream. Note that the dynamic node memory still needs to be obtained following the serialized order in the memory process. For all methods, the node memory and cached mails are stored in the main memory and transferred between CPU and GPU in every training iteration.

\subsection{Convergence}
We first evaluate the convergence of DistTGL by comparing the validation accuracy after different numbers of training iterations and the testing accuracy for the final model. 

We start with the performance of epoch parallelism on the Wikipedia, Reddit, Flights, and MOOC datasets, as the largest batch sizes on these datasets do not allow mini-batch parallelism. Figure~\ref{fig:epoch+memory}(a) shows the convergence curves of applying 1 (as the baseline), 2, 4, and 8 epoch parallelism. When $j=2$, we observe more than $2\times$ speedup for the number of training iterations before reaching 70\%, 80\%, and 90\% of the best validation accuracy on all four datasets, especially on the Flights datasets where the final test accuracy is even higher than the baseline. We believe that the super-linear scalability is due to the larger global negative batch size, where we observe similar convergence speed improvement when we increase the number of negative samples during training for the baseline. Unfortunately, increasing the number of negative samples cannot be used to speedup the convergence as the computation complexity is linear with the number of root nodes. When $j=4$, epoch parallelism still manages to achieve linear speedup except on the Flights dataset with the most number of unique edges~\cite{poursafaei2022towards}. When $j=8$, epoch parallelism leads to significant test accuracy drop and non-linear speedup. The sub-linear scalability for epoch parallelism when $j$ is large is expected as it trains on the same positive nodes consecutively in multiple iterations, leading to increased variance in the mini-batch gradients. 

\begin{figure}[t]
    \centering
    \input{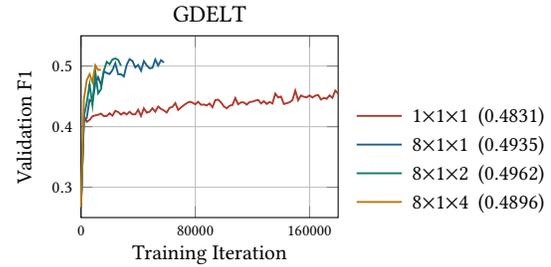}
    \caption{Convergence of DistTGL on the GDELT datasets. The test F1-Micro is shown between parentheses in the legend.}
    \label{fig:large}
\end{figure}

\begin{figure*}[t]
    \centering
    \input{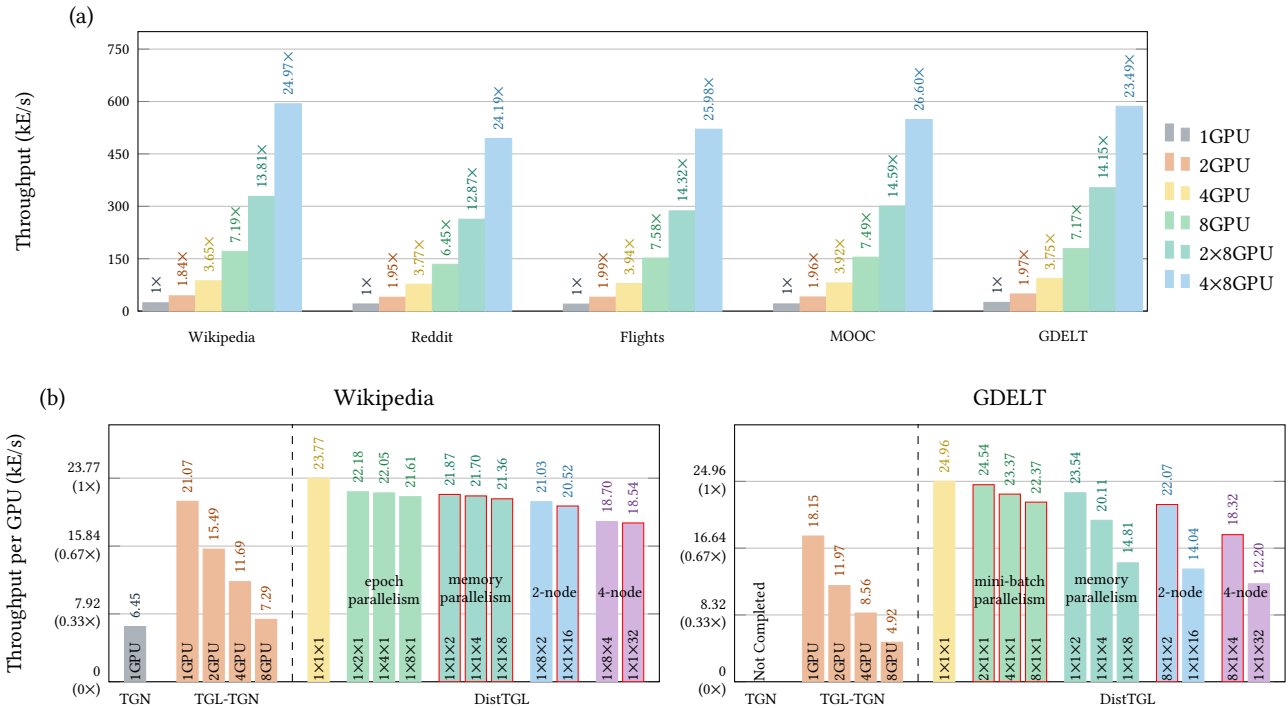}
    \caption{(a) Training throughput of DistTGL. We show the parallel training strategies with the best accuracy (memory parallelism on the four small datasets and mini-batch parallelism on the two large datasets on each node) for each dataset. The bars with red frame denote the optimal training configuration on different number of GPUs. (b) Training throughput per GPU of DistTGL compared with TGN and TGL-TGN on the Wikipedia and GDELT datasets.}
    \label{fig:throughput}
\end{figure*}

Then, on the same four datasets, we fix $j\times k=8$ and evaluate the convergence with different memory parallelism. Figure~\ref{fig:epoch+memory}(b) shows the convergence curves of different epoch and memory parallelism. 
Compared with epoch parallelism ($1\times8\times1$), memory parallelism achieves both better validation accuracy and notably better test accuracy due to better gradient estimation in each mini-batch. In general, the larger the memory parallelism $k$ is, the better the test MRR. The training configuration with the largest $k=8$ achieves linear speedup in convergence compared with the single GPU baseline with only an average of 0.004 drop in test MRR. Figure~\ref{fig:heat} shows the test MRR and the number of training iterations to reach the best validation MRR of different training configurations when $i=1$ and $j\times k\leq32$. The experiment results agree with our strategy for optimal training configuration, where we prioritize memory parallelism over epoch parallelism within the hardware limit.

For the GDELT dataset, we verify that the largest batch size without accuracy loss is larger than the capacity of one machine (see Figure~\ref{fig:motivation}(a)), which also agrees with previous work~\cite{tgl}. Hence we follow our optimal training configuration choosing policy and prioritize mini-batch parallelism. Figure~\ref{fig:large} shows the convergence of DistTGL on the GDELT datasets. The single GPU baseline $1\times1\times1$ converges very slowly. Increasing the learning rate can speedup the convergence to some extent but will also lower the accuracy. By contrast, mini-batch parallelism $8\times1\times1$ enjoys the benefit of larger batch size and achieves super-linear speedup. To further speedup on more trainers, we need to use memory parallelism to solve the massive communication overhead across machines. On multiple machines, the combination of memory parallelism and mini-batch parallelism achieves satisfying convergence speedup with the highest test accuracy. 

\subsection{Training Throughput}

We evaluate the training throughput of DistTGL on up to four 8-GPU machines. We do not test on more machines as the training time on the largest GDELT dataset is already less than 30 minutes on four machines while it only takes a few minutes to train on the smaller datasets. Figure~\ref{fig:throughput}(a) shows the training throughput and the speedup compared with the single GPU baseline of the optimal training configuration on 2, 4, and 8 GPUs on a single machine, 16 GPUs on two machines, and 32 GPUs on four machines. On 8/32 GPUs on 1/4 machines, DistTGL achieves close to linear speedup averaging 7.27/25.08$\times$, respectively. In terms of absolute throughput, the training throughput on the Reddit and Flights datasets is around 10\% slower than the other datasets due to the larger amount of writes to the node memory and cached mails. Since DistTGL only applies memory parallelism across machines, the memory operations are evenly distributed to each machine. There is no cross-machine traffic besides the synchronization of model weights, leading to a balanced workloads in each trainer. Due to the small TGNN models with only a few megabytes of weights, DistTGL also achieves near-linear speedup scaling on distributed systems.

We also compare the performance of DistTGL with the vanilla single GPU implementation TGN~\cite{tgn} and its optimized version TGL-TGN~\cite{tgl} that supports single-machine multiple-GPU. Figure~\ref{fig:throughput}(b) shows the training throughput per GPU of the two baseline methods and DistTGL in different training configurations on the Wikipedia and GDELT datasets. On the GDELT dataset, TGN does not finish training in 10 hours. DistTGL with the optimal training configurations (memory parallelism on the Wikipedia dataset and a combination of mini-batch and memory parallelism on the GDELT dataset) significantly outperform TGN and TGL. On 2, 4, and 8 GPUs, DistTGL achieves an average of 1.24$\times$, 1.91$\times$, and $2.93\times$ improvement, respectively, compared with TGL. The $1\times1\times1$ single GPU implementation of DistTGL is also faster than TGL due to our system optimization that overlaps the read and write operations from and to node memory. On the GDELT dataset, memory parallelism does not scale linearly on 8 GPUs due to the limitation of the bandwidth between CPU and RAM, whereas the scalability is notably better on multiple machines. 
\section{Conclusion}
In this work, we propose DistTGL, an M-TGNN training framework for large-scale distributed M-TGNN training.
DistTGL addressed the accuracy loss issue and communication overhead challenges by adopting three improvements of an enhanced model, a novel training algorithm, and an optimized system. 
Compared with state-of-the-art TGNN framework TGL~\cite{tgl}, DistTGL not only outperforms TGL both in convergence rate and training throughput on a single machine but also extends M-TGNN training to distributed systems.
We will open-source DistTGL and all datasets used in this work upon acceptance of this work.

\bibliographystyle{ACM-Reference-Format}
\bibliography{cite}

\end{document}